# Hybrid stemming for Persian

Adel Rahimi

*Imam Khomeini International University*
*Iran, Qazvin*
`Rahimi.adel@gmail.com`

**Abstract—** Stemming has been an influential part in Information retrieval and search engines. There have been tremendous endeavours in making stemmer that are both efficient and accurate. Stemmers can have three method in stemming, Dictionary based stemmer, statistical-based stemmers, and rule-based stemmers. This paper aims at building a hybrid stemmer that uses both Dictionary based method and rule-based method for stemming. This ultimately helps the efficacy and accurateness of the stemmer.

## I. INTRODUCTION

Stemming is the process reducing morphological bounding of a given word to their stem, root or base. It can be used to determine the morphologically similar words. Stemming has been in focus for several decades because of its use and importance. Stemmers are widely used in fields such as Information Retrieval.

Stemmers are categorized into three subcategories by their stemming method. Dictionary-based stemmers (or Lookup Table stemmers), statistical-based stemmers and Rule-based stemmer (affix removal stemmers).

Dictionary-based stemmers have every single word and its stem stored in a database and each time a word is being stemmed the program looks up the word in the dictionary and backs the result from it. This method clearly consumes high resources and the database should be updated manually for new words.

Statistical-based stemmers use statistical methods for stemming based on different corpora. This approach is based on creating rules for word formation based on statistical methods. Some methodologies formed on this basis are: frequency count, N-gram (Mayfield and McNamee, 2003), Link Analysis (Bacchin et al., 2002), and Hidden Markov Models (Melucci and Orio, 2003). This method does not rely on linguistic rules and does not need any trainings and therefore is independent of morphological rules.

Affix-removal stemmers are completely dependent on language rules. In this approach the affixes are first recognized and then added to the algorithm in order to be removed. Porter stemmer (Porter, 1980) is a perfect instance of this approach.

Persian is an Indo-European rich morphological language its writing system is Right-to-left and the affixal system consisting mainly of suffixes and few prefixes. There are more than 30 thirty suffixes in Persian language which some of the suffixes are from Arabic and others Persian roots. Since stemming plural forms in Persian can be tricky, because of numerous suffixes and different roots, and useful at the same time we mostly focus on stemming plural noun forms in Persian.

Persian Plural suffixes follow as:

TABLE I
PERSIAN PLURAL MAKING SUFFIXES

| Persian plural suffixes | |
|---|---|
| Suffix | Example |
| ‑ان | گیاهان<br>(giya-hän)<br>(trees) |
| ‑ون | روحانیون<br>(row-häni-yoon)<br>(Clergies) |
| ‑ین | مسلمین<br>(Mos-lemin)<br>(Muslims) |
| ‑ات | ملاحظات<br>(mola-hezät)<br>(Considerations) |
| ‑ها | گل ها<br>(Gol-hä)<br>(flowers) |

Yet there is another way of making plurals in Persian which is derived from Arabic is called Mokkasar (مکسر) that literally means broken. In Mokkasar approach there are set of nouns that are fixed and their equivalent is fixed as well. So trying to pluralize a noun instead of using suffixes one should use the irregular form of the word.

TABLE II
PERSIAN MOKASSAR PLURAL FORMS

| Persian irregular plural form | |
|---|---|
| word | plural |
| اثر<br>(asar)<br>(writing) | آثار<br>(äsär)<br>(writings) |
| اسم<br>(esm)<br>(name) | اسامی<br>(asämi)<br>(names) |
| جزیره<br>(jazireh)<br>(Island) | جزایر<br>(jazäyer)<br>(Islands) |
| حادثه<br>(haadeseh)<br>(Accident) | حوادث<br>(havädes)<br>(Accidents) |

| قانون | قوانین |
|---|---|
| (ghänoon) | (ghavänin) |
| (Rule) | (Rules) |

## II. PROPOSED ALGORITHM

### I. General information on the algorithm

The algorithm proposed in this paper uses both look-up table and Affix removal. The affix removal stage starts with removing 11 (eleven) suffixes that are used in word formation in Persian. This 11 suffixes include:

- ها (pronounced Ha)
- ی (Ye)
- یی (double Ye )
- ش (Shin)
- ت (Te)
- م (Mim)
- تر (pronounced Tar)
- ترین (pronounced Tarin)
- ان (Alef Nun)
- ات (Alef Te)
- ًO (Arabic fathatan, Character code:064b)

This solely is not definitely enough and cannot provide accurate results. Because firstly there is Mokassar plural form which although is low in number of entries used tremendously. As for an example we tested on VOA corpus (Jon Safari) the total number of words in the corpus is roughly 8 million and the previous examples for Mokassar plural forms result as:

TABLE III
PERSIAN MOKASSAR WORDCOUNT IN VOA CORPUS

| Word | Word count |
|---|---|
| آثار | 559 |
| اسامی | 283 |
| جزایر | 30 |
| حوادث | 80 |
| قوانین | 174 |
|  | 1126 |

A total number of 1126 entries from the 5 random words selected, roughly consists 0.01% of the corpus. Thus a database of Mokassar wordlist has been proposed. This list contains most of the frequent Mokassar words and their plural form.

The second problem arises when the stemmer tries to stem words that end with one of the suffixes above but the suffix is not part of their morphological word formation. Examples are:

TABLE IV
INTERVENING WORDS IN PERSIAN

| Intervening Words | |
|---|---|
| suffix | Examples |
| ون | ستون |
|  | هدفون |
|  | تلویزیون |
| ین | عین |
|  | دین |
|  | پایین |
| ات | اثبات |
|  | ادات |
| ان | آبادان |
|  | آبان |
|  | خان |

The Intervening database is a 128K word database of words that have the same ending as plural making suffixes nevertheless they are not plural.

### II. How does the algorithm work?

The algorithm starts with checking the word ending. There would be 5 possibilities.

- ون
- ین
- ات
- ان
- ها

In the next step the algorithm checks if the given word exist in either Intervening wordlist or Mokassar wordlist, if the given word exist in either the stem will be looked up from the dictionary and printed as the result otherwise it will go through the affix removal phase in which the stemmer removes the following suffixes if exist:

- ها (pronounced Ha)
- ی (Ye)
- یی (double Ye )
- ش (Shin)
- ت (Te)
- م (Mim)
- تر (pronounced Tar)
- ترین (pronounced Tarin)
- ان (Alef Nun)
- ات (Alef Te)
- ًO (Arabic fathatan, Character code:064b)
- ون
- ین

The overall process of stemming for a given word can be summarized as:

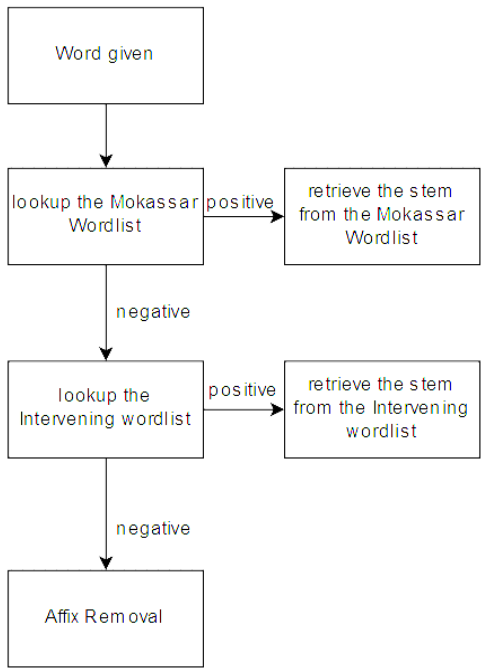

Fig. 1. Overall procedure of the algorithm

III. EVALUATION OF THE ALGORITHM

This algorithm can increase the precision of the stemming process as compared to affix-removal-only stemmers and also decreasing the database size compared to look-up-only stemmers as the database is only limited to certain specific words.

We tested this method on VOA corpus of Farsi (Jon safari) for 100 random selected words and the results are:

TABLE V
SENSITIVITY AND SPECIFICITY OF THE METHOD

| Performance of the method | |
|---|---|
| **Positive** | **Negative** |
| True Positive(TP): 13 | True Negative(TN): 83 |
| False Positive(FP): 3 | False Negative(FN): 0 |

This indicates that the total number of words detected and the affix removal stage was done on them is 13 and the number of words that were marked as either "Intervening" or "Mokassar" was 83. The number of words that actually didn't need the affix removing part and the suffix was part of their morphological word formation is 3. Finally the number of words that they should have been undergone the affix removal stage but instead were ignored is 0.

Though the rate for False Negative (FN) can be affected by Semantics. Homonyms, for example, can easily alter this factor. For instance the word سلامت can have two meanings: سلام تو (salam-e to, your greetings), and سلامت(salämat, well-being) which can either be stemmed or included in the Intervening list.

IV. CONCLUSIONS

This method provides accuracy on stemming process with using both Intervening and Mokassar wordlist. This method provides 97% accuracy on stemming process and the remaining can be achieved by considering the following factors:
    a) the comprehensiveness of the databases
    b) the semantic context

The databases can nearly be perfect and include most of the frequent words used in Persian, the semantic context on the other hand is much more complex and requires more work on Language Understanding.

On conclusion this method can provide us with accurate results for the price of using less resources.